# Computational Models for SA, RA, PC Afferent to Reproduce Neural Responses to Dynamic Stimulus Using FEM Analysis and a Leaky Integrate-and-Fire Model

Hiroki Ishizuka, Shoki Kitaguchi, Masashi Nakatani, Hidenori Yoshimura, Fusao Shimokawa

## Abstract

Tactile afferents such as (RA), and Pacinian (PC) afferents that respond to external stimuli enable complicated actions such as grasping, stroking and identifying an object. To understand the tactile sensation induced by these actions deeply, the activities of the tactile afferents need to be revealed. For this purpose, we develop a computational model for each tactile afferent for vibration stimuli, combining finite element analysis finite element method (FEM) analysis and a leaky integrate-and-fire model that represents the neural characteristics. This computational model can easily estimate the neural activities of the tactile afferents without measuring biological data. Skin deformation calculated using FEM analysis is substituted into the integrate-and-fire model as current input to calculate the membrane potential of each tactile afferent. We optimized parameters in the integrate-and-fire models using reported biological data. Then, we calculated the responses of the numerical models to sinusoidal, diharmonic, and white-noise-like mechanical stimuli to validate the proposed numerical models. From the result, the computational models well reproduced the neural responses to vibration stimuli such as sinusoidal, diharmonic, and noise stimuli and compare favorably with the similar computational models that can simulate the responses to vibration stimuli.

## Introduction

Our tactile senses can perceive not only the shape and material of an object but also the texture of an object, enabling us to perform actions such as grasping, stroking, and identifying an object. Tactile afferents located in the skin that respond to external stimuli enable these complicated actions. Usually, sensory evaluations are performed to interpret the tactile sensation induced by these actions. To understand the perceived tactile sensation quantitatively, it is necessary to reveal the relationship between the skin deformation induced by an object and the activities of tactile afferents in the skin.

 Of note, there are two possible methods to understand how the tactile afferents are activated: the first is to directly measure the action potential of tactile afferents by inserting electrodes into nerve fibers [1-3]. Electrophysiological recordings have been performed to measure the responses of tactile afferents to stimuli on skin. For example, Talbot et al. measured the responses of tactile afferents to vibration and indentation stimuli in Macaca mulatta [4]. Muniak et al. vibrated the fingers of Macaca mulatta with a probe and measured the relationship of the frequency and amplitude of the probe with and firings of SA, RA, PC afferents [5]. The authors also contrasted the intensity of the stimuli perceived by human subjects with the results obtained by examining Macaca mulatta in an attempt to reveal the relationship between human tactile perception and activities of tactile afferents. Yoshioka et al. measured activities of tactile afferents when stroking gratings of different shapes and found that the activities of the tactile afferents varied depending on the shape of the grating [6]. Weber et al. measured the activities of the tactile afferents when tracing various textures [7]. Although this method can bring important neurophysiological data, the insertion of electrodes and the restraint of the subjects and experimenters, which placed heavy mental and physical burdens on both subjects and experimenters, are necessary. In addition, specimens such as Macaca mulatta are difficult to maintain for the duration of the experiment, and it is difficult for the experimenters to perform experimental techniques. The second method is to develop numerical models of tactile afferents and analytically evaluate the activities of the tactile afferents when a stimulus is applied to the skin. This method reproduces the behaviors of the tactile afferents by adjusting the parameters in neural dynamic models, such as an integrate-and-fire model and a Hodgkin–Huxley model [8]. The parameters in the numerical models are often tuned by comparing the responses of the numerical models with the observed responses. This method makes it possible to estimate the behavior of the tactile afferents in response to a given stimulus simply by using a computer. Dong et al. constructed numerical models that calculate the activities of the tactile afferents by substituting the displacement of the skin as an input into a generalized integrate-and-fire model [9, 10]. The responses to sinusoidal stimuli, diharmonic stimuli, and noise stimuli were calculated using the numerical models for the tactile afferents, and the number and the timing of firings were compared with the observed responses to confirm that the behaviors of the tactile afferents were reproduced using the numerical models. Saal et al. proposed numerical models that reproduce the activities of the tactile afferents by adding a noise term and filter processing to Dong et al.'s model [11]. This revised numerical model can reproduce the real-world responses to ramp-and-hold stimuli as well as vibration stimuli. Ouyang et al. applied an electromechanical circuit model to reproduce real-world responses of tactile afferents [12] and confirmed that the responses of SA, RA, or PC afferents to skin indentation could be reproduced. Later, Ouyang et al. extended these numerical models. The extended models were able to incorporate population responses of many tactile afferents [13]. The authors also revealed that the extended models can reproduce real-world responses of tactile afferents to dot textures, embossed letters and curved surfaces. Although Ouyang et al.'s models can be implemented in Python and reproduce the tendency of firing rate changes, they do not adequately reproduce the response of the tactile afferents to vibration stimuli. The numerical models described above mainly reflect the activities of tactile afferents, and only take into account the component of skin deformation in the vertical direction. To expand the range of applications, it is necessary to combine numerical models of tactile afferents with a numerical method that can include cases such as complicated skin deformation.

In this study, we propose computational models that substitute skin deformation derived from finite element method (FEM) analysis into the integral firing model. FEM analysis is a numerical method of dividing an object into small elements and calculating the deformation of the entire object from the deformation of these elements; this method can be used to analyze the deformation of elastic and viscoelastic bodies. Many researchers have simulated skin deformation, and the results closely reflected actual skin deformation [14-21]. In addition, the existing studies suggest that strain obtained by numerical analysis tends to adequately reflect the activities of SA and RA afferents [22-24]. In fact, Lesniak and Gerling calculated the number of firings of SA afferents by calculating the deformation with the finger cross-section model and substituting the strain energy density (SED) into the integral firing model [25]. The authors found that the number of firings was close to the observed result. Hamsaki and Iwamoto constructed a numerical model to derive the relationship between von Mises stress and firing rate and found that it is possible to estimate the activity of SA afferents induced by stroking based on this relationship [26]. Later, Hamsaki and Yamada utilized a leaky integrate-and-fire model to estimate the firings of SA afferents from the calculated SED, and the researchers well predicted the number of firings and the timing of firings [27]. Wei et al developed an integrated numerical model using FEM analysis and Izhikevich neural dynamic model to predict the cutaneous neural dynamics during active touch [28]. So far, only the estimation of the activities of SA and RA afferents has been made using the FEM analysis and the neural dynamic models. In addition, the mentioned models have mainly been used to calculate responses to static pressure and tactile motion, and no attempt has been made to reproduce the response of each tactile afferent to dynamic stimuli, such as vibration, as was done in the experiment by Muniak et al [5], although the responses of tactile afferents to vibration are important for understanding texture recognition. Therefore, we constructed neural dynamic models to derive the response of each tactile afferent to vibration stimuli from stresses obtained by FEM analysis using the leaky integrate-and-fire model. The proposed neural dynamics models are based on Saal et al.'s model [11] and uses finite impulse response (FIR) filters that can be computed in a short time. Thanks for the simpleness of the structure, this model is expected to be implemented not only for FEM analysis, as was the case in this study, but also for processing tactile information in robots. To develop the computational models, the observed responses of SA, RA, and PC afferents to sinusoidal stimuli were first utilized in the optimization process, and the undetermined parameters in the neural dynamic models were optimized based on these data. After that, we evaluated the firings of the tactile afferents to various vibrations using these models to validate the entire model.

## 2. Overview of the proposed computational model

The proposed computational models consist of two calculations: the calculation of skin deformation and the calculation of the activities of each tactile afferent (Fig. 1). The models for calculating the activities of the tactile afferents contain undetermined parameters, and the parameters were optimized using observed biological data.

We employed a two-dimensional finite element method (FEM) model to calculate deformation more accurately since the FEM can include information about the skin shape and its physical characteristics. The calculated von Mises stress, which is related to skin deformation, was converted to a membrane current using a stress-to-current transform model that mimics the characteristics of each tactile afferent [10, 11, 25]. The membrane potential was calculated by substituting the membrane current into the leaky integrate-and-fire model. In this study, we selected a single node in the FEM model; this node represented the center of the tactile afferent, and calculated the membrane current using the change in von Mises stress at the node to simplify the physical and numerical model of the tactile afferent.

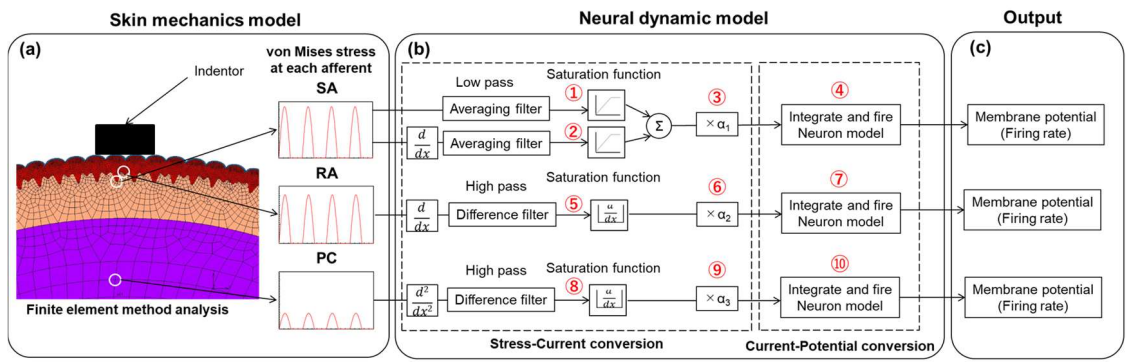

**Fig. 1 Overview of the proposed computational model. The model can be divided into two parts: skin mechanics and neural dynamics. (a) Skin deformation is calculated using FEM analysis. Von Mises stress at each tactile afferent position is sampled. (b) Von Mises stresses are converted into current to tactile afferents using FIR filters. The current is entered into the integrate-and-fire model. (c) Membrane potentials are calculated using parts (a) and (b). The models include ten unknown parameters (①-⑩). The parameters are obtained using an optimization method.**

**2.1 Model of skin mechanics**

Although the related studies have used not only two-dimensional FEM models [14-18, 20, 25, 29] but also three-dimensional FEM models [19, 21, 22, 24, 26, 27, 30, 31], we employed a two-dimensional FEM model because of its low computational cost due to the small number of elements and the ease of modeling. The shape of the FEM model used in this study was determined by referring to the dimensions of the FEM models used in existing studies [14, 17, 25, 29], and the outline of the model shape was formed using Adobe Illustrator (Adobe Inc., San Jose, CA). The FEM model consists of three layers: the epidermis, dermis, and subcutaneous tissue. Skin structures, such as fingerprints, dermal papillae, a bone, and a nail, are reproduced in the model. We imported the formed shape into Marc/Mentat (MSC Software, Santa Ana, CA) and divided the shape into meshes, as shown in Fig. 2 (a). The numbers of elements and nodes were 17815 and 36346, respectively. Deformation behavior was modeled by a plain-strain approximation. In this analysis, the iso-parametric quadrilateral element was applied. In addition, to improve the calculation tolerance of the contact condition, contacted bodies, an indentor and a fingertip were not set to have a master-slave relation, but the contact region was judged by searching all segments of both bodies (segment-to-segment). The frictional condition was Coulomb's friction law with finite slipping. The widths of the elements ranged from approximately 0.02 to 0.45 mm per side, and the elements near the fingerprints were made smaller so that they could include small deformations. In addition, the sizes of the elements inside the finger, which were not used for contact determination, were made larger to reduce the computational cost. Parameters, such as Young's modulus and Poisson's ratio of the skin, are shown in Table 1, referring to existing studies [16, 18, 30]. The correlation between the activity of the SA afferent and strain energy density (SED) in the skin has been suggested [22, 25]. In this study, von Mises stress, which is correlated with SED, was utilized to determine the input current for the leaky integrate-and-fire model. We clarified whether the behaviors of SA, RA, and PC afferents can be reproduced using von Mises stress. The von Mises stress $\sigma_{Mises}$ is shown in Eq. (1).

$$\sigma_{Mises} = \sqrt{\frac{1}{2}\left\{(\sigma_{xx} - \sigma_{yy})^2 + (\sigma_{yy} - \sigma_{zz})^2 + (\sigma_{zz} - \sigma_{xx})^2 + 6(\tau_{xy}^2 + \tau_{yz}^2 + \tau_{zx}^2)\right\}} \qquad (1)$$

where $\sigma_{xx}$, $\sigma_{yy}$, and $\sigma_{zz}$ are normal stresses and $\tau_{xy}$, $\tau_{yz}$, and $\tau_{zx}$ are shear stresses. The von Mises stress is equivalent stress and can incorporate the effects of stress states in both normal and tangential directions. In this study, we selected one node which was considered to be located on the center of each tactile afferent as shown in the upper-right area of Fig. 2 (a). We selected total three nodes in this study to calculate membrane current and firing rates for SA, RA and PC afferents. The simulation for skin mechanics was performed using Marc/Mentat. The time increment for FEM analysis was set to 0.5 ms.

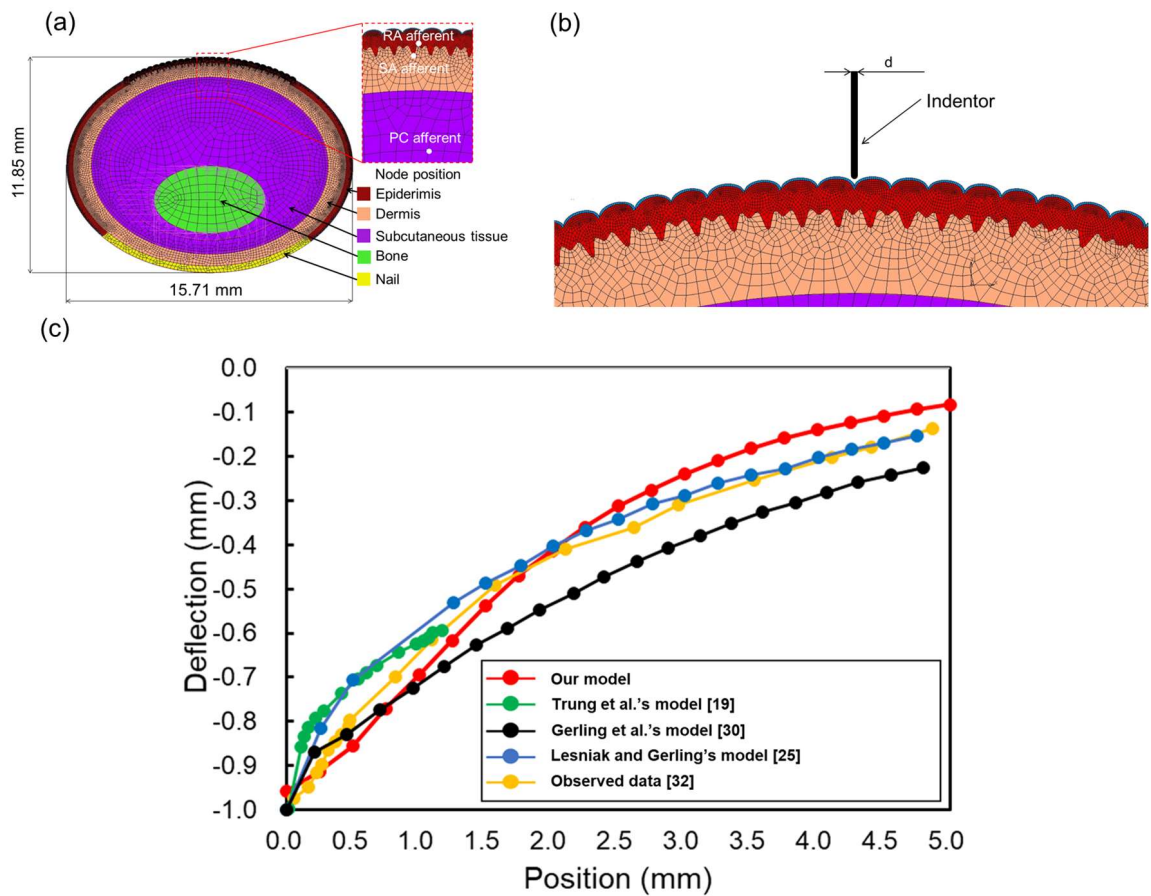

**Fig. 2 The calculated skin surface deflection was close to the value of the observed skin surface deflection, similar to the results of related studies. (a) Prepared 2D FEM model of a human finger. The model consists of the epidermis, the dermis, subcutaneous tissue, bones, and nails. The design was determined by referring to related studies [14, 17, 25, 29].**

**The enlarged figure (upper-right) shows nodes where von Mises stresses were sampled. (b) Image of FEM analysis results. In the first FEM analysis, the simulation condition is same as the related study [32]. Displacement of 1 mm was applied to the skin model using an indentor with a diameter (d) of 50 μm. In the second FEM analysis, we simulated sinusoidal vibration stimuli [5]. The displacement and frequency of the indentor were varied and the diameter (d) of the indentor was 1 mm. In the third FEM analysis, diharmonic and noise stimuli were applied. (c) In the first experiment, the maximum skin surface deflection was approximately 1 mm and the skin surface deflection was gradually decreased.**

Table 1 List of material parameters.

| Layer | Elastic modulus (MPa) | Poison ratio |
|---|---|---|
| Stratum corneum | 2.000 | 0.30 |
| Epiderimis | 2.000 | 0.30 |
| Dermis | 0.050 | 0.48 |
| Subcutaneous tissue | 0.024 | 0.40 |
| Bone | $1.700 \times 10^4$ | 0.30 |
| Nail | $1.700 \times 10^2$ | 0.30 |

**2.2 Neural dynamic model**

We calculated the membrane potential and the resulting firing rate for each tactile afferent using the stress-to-current transform models and the leaky integrate-and-fire model [8, 10, 11, 25].

It is known that the SA afferent is not sensitive to high-frequency vibration, and RA and PC afferents are sensitive to high-frequency vibration [1, 5]. To reproduce these characteristics, von Mises stress, its derivative, and its second-order derivative were utilized by referring to the known characteristics of the tactile afferents and were filtered before being incorporated into the stress-to-current transform model.

$$\sigma_{filtered\ SA1}(t) = \sum_{n=-m_2}^{m_1} \frac{|\sigma_{SA}(t+ndt)|}{m_1 + m_2 + 1} \quad (1)$$

$$\sigma_{filtered\ SA2}(t) = \sum_{n=-m_4}^{m_3} \frac{|\sigma'_{SA}(t+n\ )|}{m_3 + m_4 + 1} \quad (2)$$

$$\sigma_{filtered\ RA}(t) = |\sigma'_{RA}(t) - \sigma'_{RA}(t-dt)| \quad (3)$$

$$\sigma_{filtered\ PC}(t) = |\sigma''_{PC}(t) - \sigma''_{PC}(t-dt)| \quad (4)$$

where $t$ is time; $dt$ is the time increment; $\sigma_{filtered\ SA1}(t)$, $\sigma_{filtered\ SA2}(t)$, $\sigma_{filtered\ RA}(t)$, and $\sigma_{filtered\ PC}(t)$ are filtered von Mises stresses; $\sigma_{SA}(t)$, $\sigma_{RA}(t)$, and $\sigma_{PC}(t)$ are von Mises stresses sampled at a node for each tactile afferent; and $m_1$, $m_2$, $m_3$, and $m_4$ are parameters for the filter width. An averaging filter, that acts as a low-pass filter, was applied to the von Mises stress and the derivative at the SA afferent. A difference filter was used to emphasize the change in the derivative and the second-order derivative of the von Mises stresses at RA and PC afferents. The values of $m_1$, $m_2$, $m_3$, and $m_4$ were 9, 9, 9, and 8, respectively.

Although the correlations between the responses of tactile afferents and von Mises stress are not perfectly revealed, we constructed stress-to-current transform models, referring to the existing numerical models for tactile afferents [10, 11]. The stress-to-current transform models are expressed as follows:

$$I_{SA}(t) = \alpha_1 \left( \frac{|\sigma_{filtered\ SA1}(t)|}{a_1 + |\sigma_{filtered\ SA1}(t)|} + \frac{|\sigma_{filtered\ SA2}(t)|}{a_2 + |\sigma_{filtered\ SA2}(t)|} \right) \quad (4)$$

$$I_{RA}(t) = \alpha_2 \left( \frac{|\sigma_{filtered\ RA}(t)|}{a_3 + |\sigma_{filtered\ RA}(t)|} \right) \quad (5)$$

$$I_{PC}(t) = \alpha_3 \left( \frac{|\sigma_{filtered\ PC}(t)|}{a_4 + |\sigma_{filtered\ PC}(t)|} \right) \quad (6)$$

where $I_{SA}(t)$, $I_{RA}(t)$, and $I_{PC}(t)$ are the membrane currents and $\alpha_1$, $\alpha_2$, and $\alpha_3$, $a_1$, $a_2$, $a_3$, and $a_4$ are undetermined parameters in the stress-to-current transform models. We noted that the units of $\alpha_1$, $\alpha_2$, and $\alpha_3$ are mA. The filtered von Mises stress and filtered derivative are incorporated into the stress-to-current transform model for SA afferents. On the other hand, only the filtered derivative of von Mises stress or the filtered second derivative of von Mises stress was incorporated into the transform model for RA afferents or PC afferents, respectively. The observed firing rate was saturated when the stimulus intensity exceeded a certain level [5]. To represent the characteristics, the stress-to-current transform models can saturate the resulting current, owing to their structure.

By substituting the current value obtained by the transform model into the leaky integrate-and-fire model, the membrane potential and firing rate were obtained.

$$\frac{du}{dt} = -\frac{1}{\tau_m}(u(t) - u_{rest}) + \frac{R_m}{\tau_m} I(t) \quad (7)$$

where $u(t)$ is the membrane potential, $\tau_m$ is the time constant that is tuned for each tactile afferent, $u_{rest}$ is the resting membrane potential, $R_m$ is the resistance, and $I(t)$ is the current derived from the stress-to-current transform model (Eq. (4)-(6)). $u_{rest}$ and the initial membrane potential were set to -65 mV. When the membrane potential exceeded a threshold, a spike was assumed to have occurred, the time was recorded, and the potential was returned to $u_r$. The threshold for SA and the thresholds for RA and PC afferents were -50 mV and -55 mV, respectively, in consideration of the sensitivities of tactile afferents, and $u_r$ is -65 mV. Of note, $I(t)$ was not inputted to reproduce the behavior of the refractory period for τ. τ for RA and PC afferents was 0.5 ms, and τ for SA afferents was 1.0 ms. The calculation and the optimization for the neural dynamic model was performed using Python. The time increment was set to 0.5 ms. We considered $\frac{R_m}{\tau_m}\alpha_1$, $\frac{R_m}{\tau_m}\alpha_2$, and $\frac{R_m}{\tau_m}\alpha_3$ to be $\alpha'_1$, $\alpha'_2$, and $\alpha'_3$, derived from Eqs. (4)-(7), for ease of the optimization.

## 3. Analysis

We validated the skin mechanics model and neural dynamic model as follows. First, we validated the two-dimensional FEM model of a cross-section of a fingertip, comparing the calculated deformation with the observed result [32] and predicted results reported by other groups [19, 25, 30]. Then, we optimized the parameters for each tactile afferent in the neural dynamic model against the observed firing rates in response to sinusoidal stimuli [5]. Finally, we calculated the responses of the optimized computational models to diharmonic and noise stimuli and evaluated the validation of the proposed computational models.

### 3.1 Skin deflection validation

We applied the parameters, shown in Table 1, to the skin mechanics model and compared the deflection obtained by FEM analysis with the observed deflection [32]. Fig. 2 (a, b) shows the FEM model. In the FEM analysis, a probe with a diameter of 50 μm was applied vertically to the fingerpad to indent it for 1 s. The friction coefficient was 0.4 since we considered the contact between skin and metal [33]. The maximum displacement of the probe was 1 mm. After the FEM analysis, the deflection of the skin surface was obtained with an interval of 0.5 mm.

### 3.2 Model optimization

We determined the parameters in the neural dynamic models by matching the predicted firing rates of tactile afferents against the observed firing rates [5]. First, we conducted FEM analysis to calculate von Mises stress induced by skin deformation. In the FEM analysis, the indentor was set to 1 mm in diameter, as shown in Fig. 2 (b), to reproduce the reported experiment; the indentor was vibrated sinusoidally for 250 ms. The amplitude and frequency of the sinusoidal vibration are summarized in Appendix A. The frequency of the sinusoidal vibration on the skin was set from 20 Hz to 300 Hz. In this analysis, the frequency of 500 Hz was not selected since there are few situations where a high-frequency vibration of several hundred Hz or higher is applied to the skin, and FEM analysis has a high computational cost. The vibration amplitude was set to about 5 μm or higher, which was in agreement with the firings of all tactile afferents observed in Muniak et al.'s results.

From the results of the FEM analysis, we obtained the temporal change in von Mises stress at one node for each tactile afferent directly under the probe, as shown in the upper-right of Fig. 2 (a), since the firing rates of the tactile afferents located under the probe were recorded in the reported experiments. The change in membrane potential and the number of firings were calculated by substituting the von Mises stress in each vibration condition with Eq. (7). We removed the results from the 100 ms at the beginning of the FEM analysis and used the remaining results in the optimization process. The undetermined parameters in the neural dynamic models were optimized using a Pareto optimal solution with a multi-objective optimization library available in Python. The Pareto optimal solution is a method that minimizes multiple objective functions. The average of the sum of the squares of the differences between the calculated and observed firing rates at each frequency was used as an objective function, and the undetermined parameters that minimized the four objective functions were searched. A multi-objective evolutionary algorithm, NSGA-II, was used as the search algorithm. The averaged values of the observed firing rates of each tactile afferent used for optimization. We searched the optimal parameters using 10,000 function evaluations. Since the Pareto optimal solution is presented with a number of candidate parameters, a combination with the smaller difference in the observed value and the closest approximation of the frequency response was selected among the candidates. We optimized $τ_m$, $a_1$, $a_2$, and $α'_1$ for the SA afferent; $τ_m$, $a_3$, and $α'_2$ for the RA afferent; and $τ_m$, $a_4$, and $α'_3$ for the PC afferent.

To evaluate the optimization results, the firing rates were calculated for each tactile afferent under the same condition as that for the optimization process. The computation resulting time was almost same as that in Muniak's measurement experiment: 245 ms for 20 Hz and 100 ms for the other frequencies.

### 3.3 Model validation

We validated the optimized parameters for the neural dynamic model for each tactile afferent and for the entire computational model by calculating the firing rate and comparing the model responses to diharmonic and band-pass noise stimuli with the reported results. First, we obtained a temporal change in von Mises stress at each tactile afferent to diharmonic and band-pass noise stimuli, using FEM analysis. The FEM model, shown in Fig. 2 (a), was applied, and the diameter of the indentor was set to 1 mm. Diharmonic stimuli applied to the indentor are shown in Appendix B. To apply band-pass noise stimuli using the indentor, we prepared Gaussian noise and filtered the noise according to the conditions shown in Appendix C, using Python. The filtered noise was modified so that the root-mean-square amplitudes became specific amplitudes. Then, the noise was transferred to Marc/Mentat. The calculated von Mises stress was substituted into the neural dynamics models to calculate the firing rate. The total computation times of FEM and the neural dynamic model were 250 ms and 245 ms, respectively.

**Result**

### 4.1 Skin deflection validation

The FEM model reproduced the skin surface deflection well, similar to the results of related studies [19, 25, 30, 32] (Fig. 2 (c)). Deflection of the skin surface was induced by skin displacement of 1 mm using the indentor. Deflection of the skin decreased as the measured position moved away from the indentor. In conclusion, the parameters that we selected were sufficient to reproduce deflection of the skin surface, and we applied these parameters in the following sections.

**4.2 Model optimization**

The optimized parameters enabled the computational models to reproduce the firing timing and the firing rate in response to sinusoidal stimuli. The obtained parameters are summarized in Table 2. The predicted spike-time rasters and the relationship between the predicted firing rate and the amplitude of sinusoidal stimulus using the parameters and Eq. (7) are shown in Fig. 3 (a, b), respectively. To obtain the rasters, we calculated the same condition five times. For comparison, Fig. 3 (b) also includes the observed responses and the responses calculated using Saal et al's model [11], and responses from Ouyang et al.'s study were replotted [13]. In Saal's model, four SA afferent models, nine RA afferent models, and four PC afferent models are available. We calculated the firing rate using MATLAB and Simulink (MathWorks, Inc., Natick, MA), averaged the predicted firing rates using the same type of afferent model, and showed the averaged firing rates in Fig. 3 (b). The computation times were 250 ms for 20 Hz and 100 ms for the other frequencies.

      We confirmed that the computational models well predicted the responses when the indentor was in contact with the skin (Fig. 3 (a)). Although there were some discrepancies between the firing timings of the computational models and those of the actual tactile afferents, it was confirmed that the computational models fired at approximately the same timings. The threshold for the firing of the computational model for SA afferents was higher than that of actual SA afferents since the computational model did not fire when the amplitude of the sinusoidal stimulus was 35 μm.

      The calculated firing rates corresponded well with the observed firing rates (Fig. 3 (b)). The upper part of Fig. 3 (b) shows that the firing rates of the computational model for SA afferents increased with the amplitude of the sinusoidal stimulus, but the firing rate did not increase significantly depending on the frequency. In particular, the proposed computational mode for SA afferents could reproduce the tendency of cells not to fire at a high frequency of 300 Hz. This outcome was observed due to the fact that the averaging filter used in the neural dynamic model functioned as a low-pass filter. Although other models for SA afferents can reproduce the characteristics of SA afferents, the proposed computational model better reproduced the characteristics of SA afferents with respect to the firing rate. For the computational model for RA afferents, as shown in the middle of Fig. 3 (b), the firing rate increased as the amplitude and the frequency increased. The other models tended to exhibit lower firing rates compared with the

observed firing rates and did not reproduce the tendency to depend on the frequency. In addition, other models exhibited no firing at 300 Hz due to the low-pass filter. On the other hand, the proposed computational model was able to reproduce the response of RA afferents at high frequencies by using the difference filter that emphasizes the change in von Mises stress, although the firing rates at 300 Hz tended to be higher than the observed firing rates. In the case of PC afferents, the firing rate increased with increasing amplitude, as shown in the bottom of Fig. 3 (b); moreover, a tendency for the firing rate to increase with increasing frequency was confirmed. Saal et al.'s model also reproduced this tendency, but Ouyang et al.'s model did not reproduce this tendency. The firing rates calculated by the proposed computational model for PC afferents were close to the observed firing rates and reproduced the characteristics of PC afferents well. $R^2$ was almost greater than 0.8 for the regression line of all plots of each tactile afferent, indicating a strong correlation in all models to the actual results (Fig. 3 (c)). P values of the regression were less than 0.05 and significant differences were confirmed. In the case of SA afferents, there was no firing at the points where the amplitude of the sinusoidal stimulus is small. This was consistent with the fact that there was no firing observed at an amplitude of 35 μm in the spike-time rasters in Fig. 3 (a). In the case of RA afferents, the coefficients were lower only for the 300 Hz case. For PC afferents, the correlation coefficients were high in all sinusoidal frequency conditions, and the overall correlation coefficient was high. In summary, it can be said that the proposed models were well optimized and able to reproduce the response of each tactile afferent, except in some conditions.

Table 2 List of optimized parameters.

| Afferent type | $\tau_m$ (ms) | $a_1$ (Pa) | $a_2$ (Pa/ms) | $a_3$ (Pa/ms) | $a_4$ (Pa/ms$^2$) | $\alpha'_1$ (mV/ms) | $\alpha'_2$ (mV/ms) | $\alpha'_3$ (mV/ms) |
|---|---|---|---|---|---|---|---|---|
| SA afferent | 32.14 | 1926.32 | 9850.98 | | | 1.79 | 10.23 | |
| RA afferent | 456.70 | | | 17191.87 | | | 10.23 | |
| PC afferent | 639.85 | | | | 16.34 | | | 4.14 |

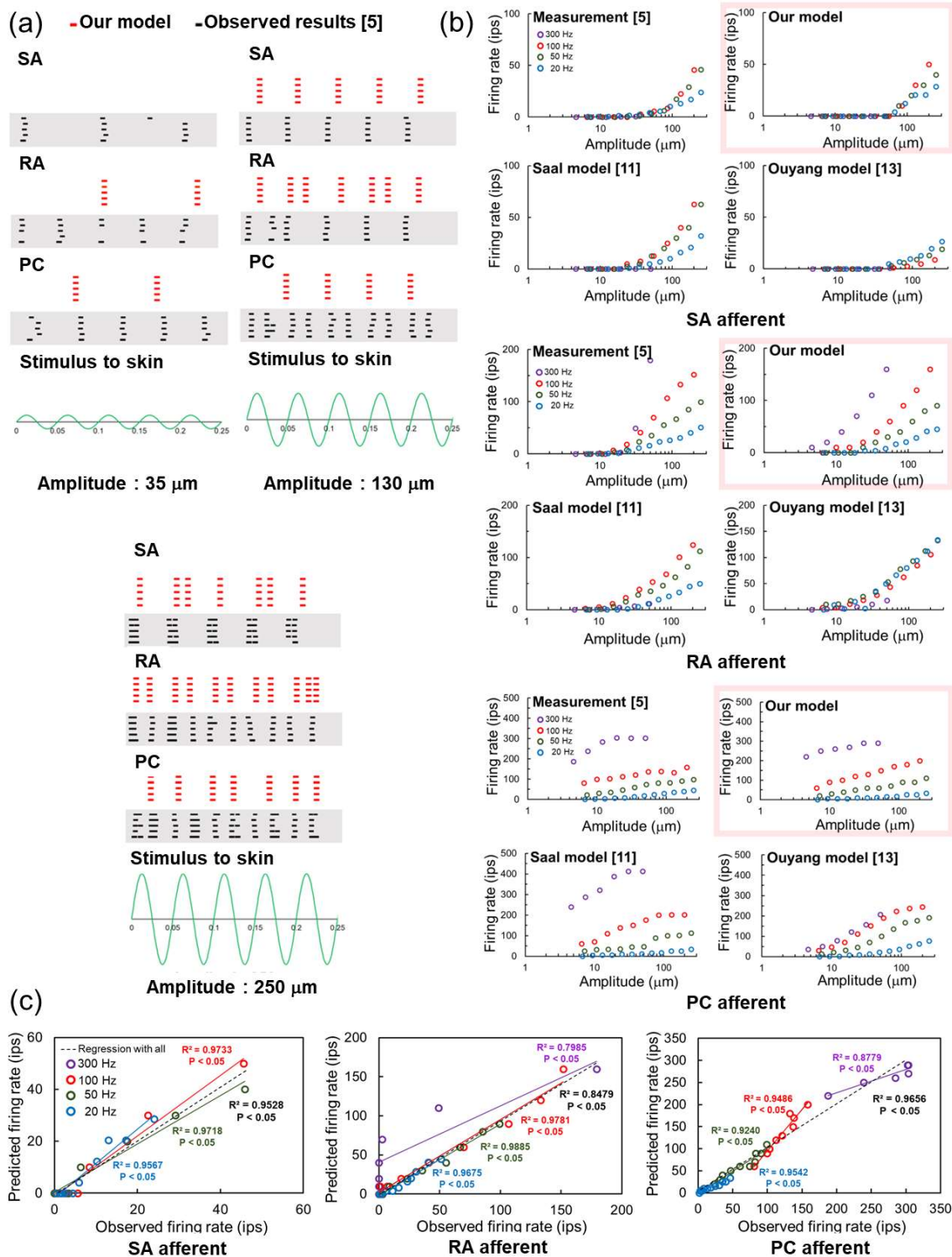

**Fig. 3 Our neural dynamic models could simulate spikes and observed firing rates reported in the literature [5]. (a)** The timings of the calculated spikes agreed well with those of the observed spikes. The frequency of sinusoidal stimuli was 20 Hz and the amplitudes were 35 μm, 130 μm, and 250 μm(c). **(b)** Our models were comparable with the conventional

**models [11, 13]. Relationships between amplitude and firing rate for each tactile afferent (upper, SA afferent; middle, RA afferent; bottom, PC afferent). (c) Correlation coefficients of regression lines are higher than 0.8 and the tendencies of firing rates were well reproduced. This indicates that our models were well trained by firing rates induced by sinusoidal stimuli. Dashed lines represent the regression lines on all plots.**

4.3 Model Validation

The proposed models suitably approximated firing rates in response to stimuli that were not used in the optimization process. Fig. 4 and Fig. 5 show the results of the diharmonic and noise stimuli, respectively. As in the previous section, the observed results and the results of other models are also shown for comparison [11, 13]. For Saal et al.'s model, the total computation time was 1 s, and we plotted the averaged firing rates using the same type of afferent model.

In the case of diharmonic vibration, the tendency for each tactile afferent corresponded well with the observed tendencies since $R^2$ was greater than 0.8 for the regression line of all plots of each tactile afferent, indicating a strong correlation in all models to the actual results (Fig. 4 (d)). P values of the regression were less than 0.05 and significant differences were confirmed. The firing rate predicted by the proposed computational model for SA afferents tended not to change with the frequency of the vibration, and the same tendency was noted in the observed result (Fig. 4 (a)). Ouyang et al.'s model exhibited the same tendency, but the firing rates of the proposed computational model were closer to the observed result. As for the computational model for RA afferents, as in the other models, little effect was noted in the difference in frequency (Fig. 4 (b)). The computational model for PC afferents was able to roughly reproduce the firing rate tendency, but the firing rate at higher frequencies was lower than the observed firing rates (Fig. 4 (c)). The tendency was similar to that of Ouynag et al.'s model, but Saal et al.'s model was able to reproduce the tendency of PC afferents better than the current model.

The firing rates for band-pass filtering of noise vibration can be predicted using the computational models especially for RA and PC afferents. $R^2$ was almost 0.8 for the regression line of all plots of RA and PC afferens (Fig. 5 (d)). P values of the regression were less than 0.05 and significant differences were confirmed.

Saal et al.'s model for SA afferents showed firing rates that are similar to the observed firing rates, but the proposed computational model and Ouyang et al.'s model for SA afferents did not show firing activity (Fig. 5 (a)). This results in the weak correlation, as shown in Fig. 5 (d).

The computational model for RA afferents was able to reproduce the tendency of the firing rate better than the other models due to the difference in the passband frequency (Fig. 5 (b)). Ouyang et al.'s model was able to reproduce the tendency to fire but could not reproduce the firing rate. For PC afferents, the proposed computational model and Saal et al.'s model well reproduced the firing rate tendencies, although the proposed computational model reproduced the firing rates at higher frequency slightly better than other models (Fig. 5 (c)). In summary, the proposed computational models were able to reproduce the tendencies of the firing rates of each tactile afferent when diharmonic or noise vibration was applied, which was not used in the optimization process. Thus, the proposed computational models are considered to be widely useable models to reproduce the responses to vibration stimuli.

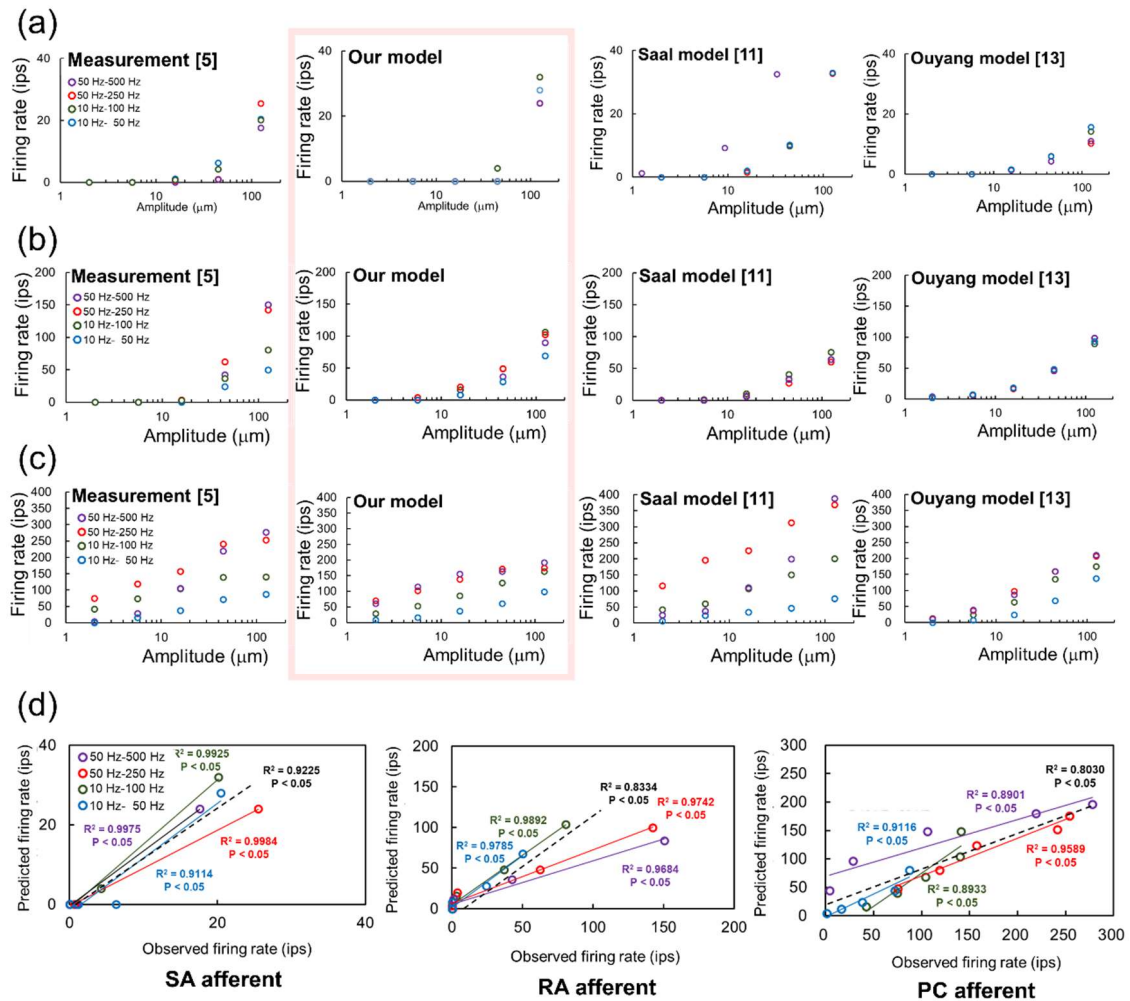

Fig. 4 Our models well represented the frequency characteristics of diharmonic stimuli, except for the difference in RA afferents induced by the frequency of the stimuli. (a) Relationships between amplitude and firing rate for SA afferents in response to diharmonic stimuli. (b) Relationships between amplitude and firing rate for RA afferents in response to diharmonic stimuli. (c) Relationship between amplitude and firing rate for PC afferents in response to diharmonic stimuli. (d) Correlation coefficients of regression lines are higher than 0.8 and the tendencies of firing rates were well reproduced. This indicates that the models well simulated the actual neural responses to diharmonic stimuli. Dashed lines represent the regression lines on all plots.

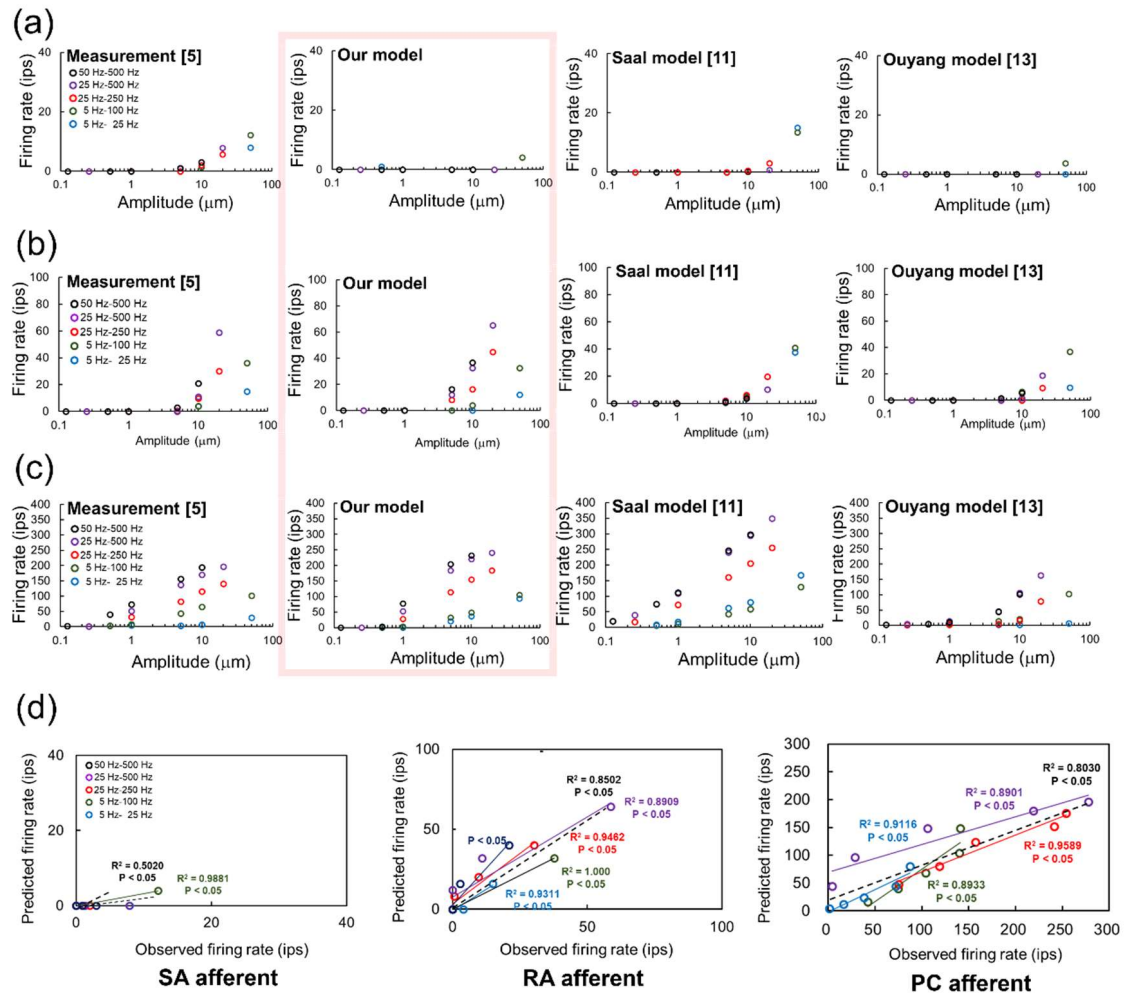

Fig. 5 Our model well predicted the firing rate, except that of SA afferents, when noise stimuli were applied. This result indicates that our model could partially simulate firing rates. (a) Relationships between amplitude and firing rate for SA afferents in response to band-pass noise stimuli. (b) Relationships between amplitude and firing rates for RA afferents in response to band-pass noise stimuli. (c) Relationship between amplitude and firing rates for PC afferents in response to band-pass noise stimuli. (d) Correlation coefficients of regression lines are higher than 0.8 except SA afferent since our model for SA afferent did not respond to band-pass noise stimuli. This indicates that the models for RA and PC afferent well simulated the actual neural responses to band-pass noise stimuli. Dashed lines represent the regression lines on all plots.

## 5. Discussion

### 5.1 Simulating tactile afferents

We developed computational models to calculate the response of each tactile afferent to skin deformation by FEM analysis, and we confirmed that the responses of SA, RA, and PC afferents to vibration stimuli could be reproduced. In the case of SA afferents, the minimum amplitudes to induce a spike were higher than the observed result. The minimum amplitudes were 67.01 μm (20 Hz), 113.60 μm (50 Hz), and 84.98 μm (100 Hz) for SA, RA, and PC afferents, respectively, in response to the sinusoidal stimuli. On the other hand, the observed minimal amplitudes were 6.70 μm (20 Hz), 23.46 μm (50 Hz), and 10.00 μm (100 Hz) for SA, RA, and PC afferents, respectively. This result may be due to a computational limitation. In this study, we calculated the firing rates using the proposed computational models with a computation time of 250 ms for 20 Hz or 100 ms for other frequencies, as was done in Muniak et al.'s experiment [5]. This means that firing rates of less than 4 ips for 20 Hz and 10 ips for other frequencies, cannot be calculated. In the case of actual measurements, the average values of the firing rate of several tactile afferents were calculated, and lower firing rates than our calculated firing rates could be obtained. For RA and PC afferents, the observed firing rates exceeded 10 ips even when the amplitude of the sinusoidal stimuli was small. We can say that the proposed computational models for RA and PC afferents could reproduce the responses to sinusoidal stimuli with small amplitudes after the optimization process. The proposed models of SA, RA, and PC afferents could reproduce the difference in response to the frequency of the sinusoidal stimulus. On the other hand, in the case of diharmonic and noise stimuli, the responses of RA and PC afferents were not partly reproduced by the model. The same tendencies were observed in Saal et al.'s model and Ouyang et al.'s models, so this outcome is considered to be a limitation of computational models. The main difference between the proposed computational models and Saal et al.'s model and Ouyang et al.'s models, is the calculation method of skin deformation using FEM analysis. Our results suggest that adding detailed skin deformation does not largely change the calculated neural responses.

In this study, we simply modeled a node in the FEM model as the center of a tactile afferent, aimed to reproduce the responses to vibration, and did not consider the size of the receptive field. Tactile mechanoreceptors, which are connected to a nerve, are distributed in three dimensions, and the distribution determines the size of the receptive field. The skin mechanics model in this study is two-dimensional and cannot perfectly address the distribution of the tactile mechanoreceptors. By increasing the number of nodes used in the calculation, the extended models could possibly reproduce the receptive field size in two dimensions. Additionally, even if we apply the neural dynamic model to three dimensional FEM model, von Mises stress and its trend to frequency will not be largely changed, and the same results can be obtained by simulation and calculation.

The biological data utilized for the optimization were not obtained from human subjects, and our computational models do not strictly model the neural responses of humans. However, the types of tactile afferents are identical among humans and Macaca mulatta. Although the minimal amplitudes and the number of the firing rates might be different between humans and Macaca mulatta due to the size of the finger, the trends of the neural responses of the tactile afferents are considered similar. Our computational models well reproduced the trends, and we expect that our computational model will model the neural responses of humans, including the trends and the numbers firing, using biological data sampled from human subjects.

## 5.2 Comparison with other models

We confirmed that the proposed computational models could reproduce the responses of the tactile afferents as well as or better than the existing computational models that can simulate the responses to vibration stimuli, although the calculation methods and the structures of the models differed among these models.

The Saal et al. and Ouyang et al.'s computational models [11, 13] aimed to reproduce the characteristics of the tactile afferents using a simplified two-dimensional skin model and neural dynamic models and included filters for processing in the frequency domain. In these computational models, as in the present study, the neural dynamic models were optimized based on the responses of the actual tactile afferents, and it was shown that the responses of the tactile afferents to vibration and lamp stimuli could be reproduced. Thus, these models are helpful in simulating the responses of tactile afferents, although numerical analysis software, such as MATLAB or Python, is required to perform filter processing. In addition, relatively high hardware performance is required.

The proposed computational model utilizes FEM analysis to calculate the skin deformation and uses Python to calculate responses of the tactile afferents from the skin deformation. Although the FEM analysis has the advantage of being able to deal with complex skin deformation, it has the disadvantage of requiring high-performance hardware. On the other hand, the neural dynamic models in the proposed computational models consist of only a simple function (saturation function), a filter (a moving filter or an edge enhancement filter) using up to 19 data points, and a differential equation (a leaky integrate-and-fire model). The advantage of this approach is that it is computationally inexpensive and can be implemented on low-performance hardware, such as inexpensive microcontrollers. The proposed computational models calculate skin deformation by FEM analysis, and it is potentially possible to include skin deformation not only in the vertical direction but also in the shear direction. Thus, it should be possible to calculate the responses of the tactile afferents due to skin deformation in the shear direction. For this purpose, it is essential to collect data from physiological experiments to modify the proposed computational model and for further optimization. Other neural dynamic models combined with FEM did not include the terms of frequency response such as a moving filter and may not well reproduce the responses to dynamic stimuli [25, 27, 28, 30].

### 5.3 Applications

As mentioned above, the proposed neural dynamic models of the tactile afferents are computationally inexpensive because it uses only a simple calculation model; this approach has an advantage in that it can be implemented in microcomputers and other devices. For example, the proposed neural dynamic models can be applied to tactile sensors and tactile sensing in robots [34-35]. For example, machine learning is often applied to identify materials from time-series signals measured by tactile sensors. Using machine learning and time-series neural spikes generated using the proposed neural dynamic models, it should be possible to implement tactile perception based on the mechanism of human tactile perception in tactile sensors and robots.

In addition, we expect that the proposed computational model can be used in the reproduction of tactile sensation in tactile displays. In many cases, tactile displays have reproduced tactile sensation by reproducing signals obtained by tactile sensors on the tactile displays [36-38]. Thus, the deformation of the skin and the activity of tactile afferents have not been taken into account. On the other hand, using the proposed model for tactile reproduction, we can calculate the effect of skin contact with an object and the activity of the tactile afferents caused by the contact. If we can determine the underlying characteristics of tactile displays that can perfectly reproduce the activities of the tactile afferents caused by skin contact with actual objects, we can make simulations of tactile sensation better reflect such sensation in the real world via tactile displays.

### 5.4 Limitations

The proposed computational models only model the tendencies of the averaged firing rates. They cannot reproduce the characteristics of individual tactile afferents as described in Muniak et al.'s study [5]. Although the calculated firing timings are close to that of the actual tactile afferents, as shown in Fig. 3 (a), the accuracy of the firing timing was not considered in this study. Thus, it is better to use the existing models if firing timing is to be considered.

This study modeled vibrations with a maximum amplitude of 250 μm and did not consider displacements beyond that value. Thus, it is not clear whether the response to the high displacement stimulus can be reproduced; Lesniak and Gerling's model attempted to address this issue[25].

In Muniak et al.'s study, frequencies of up to 600 Hz were used to apply sinusoidal stimuli to skin [5]. In this study, we applied frequencies of up to 300 Hz to optimize the undetermined parameters; this choice was made because of the computational cost of FEM analysis and the fact that RA and SA rarely respond to sinusoidal stimuli with a frequency of 600 Hz. It may not be possible to reproduce the response to high-frequency vibration using the current computational models. On the other hand, vibrations up to several hundred Hz induced by textures and generated by the tactile devices were discussed in the Application section. We consider that the response to high-frequency vibrations such as 600 Hz does not need to be considered in practice.

**Conclusion**

In this study, we confirmed whether responses of tactile afferents to vibratory stimuli could be reproduced using computational models consisting of FEM analysis and neural dynamics. A two-dimensional skin cross-sectional model was created and applied to calculate von Mises stress in the skin when a vibration stimulus was applied. The response of a tactile afferent was calculated by substituting the von Mises stress at a node assumed to be the center of a tactile afferent into the neural dynamic model. The undetermined parameters in the neural dynamic model were optimized by comparing the calculated firing rates with the observed firing rates. We confirmed that the proposed model could reproduce responses to the sinusoidal stimuli used in the optimization process and in response to diharmonic and noise stimuli. These computational models are expected to be used not only to reveal the mechanism of tactile perception that correspond to skin deformation in response to the activities of tactile afferents but also to analyze tactile sensors and tactile displays.

**Supplemental Material**

**Appendix A**

**List of parameters for sinusoidal stimuli.**

| Frequency (Hz) | Amplitude (μm) |
|---|---|
| 20 Hz | 6.71, 9.32, 12.50, 18.00, 25.000, 34.74, 48.27, 67.07, 93.19, 129.49, 179.92, 250.00 |
| 50 Hz | 7.19, 10.66, 15.81, 23.46, 34.80, 51.62, 76.58, 113.60, 168.52, 250.00 |
| 100 Hz | 6.52, 10.00, 15.34, 23.54, 36.11, 55.39, 85.98, 130.37, 200.00 |
| 300 Hz | 4.59, 7.41, 11.94, 19.24, 31.02, 50.00 |

**Appendix B**

**List of parameters for diharmonic stimulus.**

| Component 1 | | Component 2 | |
|---|---|---|---|
| Frequency (Hz) | Amplitude (μm) | Frequency (Hz) | Amplitude (μm) |
| 10 | 2.00 | 50 | 2.00 |
|  | 5.62 |  | 5.62 |
|  | 15.81 |  | 15.81 |
|  | 44.46 |  | 44.46 |
|  | 125.00 |  | 125.00 |
| 10 | 2.00 | 100 | 2.00 |
|  | 5.62 |  | 5.32 |
|  | 15.81 |  | 14.14 |
|  | 44.46 |  | 37.61 |
|  | 125.00 |  | 100.00 |
| 50 | 2.00 | 250 | 1.00 |
|  | 5.62 |  | 2.48 |
|  | 15.81 |  | 6.12 |
|  | 44.46 |  | 15.15 |
|  | 125.00 |  | 37.50 |
| 50 | 2.00 | 500 | 0.25 |
|  | 5.62 |  | 0.74 |
|  | 15.81 |  | 2.17 |
|  | 44.46 |  | 6.37 |
|  | 125.00 |  | 18.75 |

**Appendix C**

**List of parameters for band-pass noise stimuli.**

| Frequency cutoffs (Hz) | | Root-mean-square amplitude (μm) |
|---|---|---|
| Low | High | |
| 5 Hz | 25 Hz | 0.50, 1.00, 5.00, 10.00, 50.00 |
| 5 Hz | 100 Hz | 0.50, 1.00, 5.00, 10.00, 50.00 |
| 25 Hz | 250 Hz | 0.25, 1.00, 5.00, 10.00, 20.00 |
| 25 Hz | 500 Hz | 0.25, 1.00, 5.00, 10.00, 20.00 |
| 50 Hz | 500 Hz | 0.13, 0.50, 1.00, 5.00, 10.00 |